\documentclass[11pt]{article}

\usepackage[preprint]{acl}

\usepackage{times}
\usepackage{latexsym}
\usepackage[T1]{fontenc}
\usepackage[utf8]{inputenc}
\usepackage{microtype}
\usepackage{inconsolata}
\usepackage{graphicx}
\usepackage{subcaption}

\usepackage{amsmath}
\usepackage{amssymb}

\usepackage{algorithm}

\usepackage[noend]{algpseudocode}

\usepackage{xspace}
\usepackage{xcolor}

\usepackage{enumitem}

\newcommand{\ASL}{\ensuremath{\text{ASL}}\xspace}
\newcommand{\CSL}{\ensuremath{\text{CSL}}\xspace}
\newcommand{\DGS}{\ensuremath{\text{DGS}}\xspace}

\newcommand{\StoS}{\ensuremath{\text{sign}\!\to\!\text{sign}}\xspace}
\newcommand{\SStoSS}{\ensuremath{\text{sign}\!\leftrightarrow\!\text{sign}}\xspace}
\newcommand{\TS}{\ensuremath{\text{T2S}}\xspace}
\newcommand{\ST}{\ensuremath{\text{S2T}}\xspace}
\newcommand{\StwoS}{\ensuremath{\text{S2S}}\xspace}

\renewcommand{\paragraph}[1]{\vspace{3pt}\noindent\textbf{#1}}

\newcommand{\mbart}{\ensuremath{\text{\textsc{mBART}}}\xspace}
\newcommand{\soke}{\ensuremath{\text{\textsc{SOKE}}}\xspace}

\newcommand{\smplx}{\ensuremath{\text{SMPL-X}}\xspace}
\newcommand{\bos}{\ensuremath{\texttt{bos}}\xspace}
\newcommand{\eos}{\ensuremath{\texttt{eos}}\xspace}

\newcommand{\dhh}{\ensuremath{\text{DHH}}\xspace}

\newcommand{\howtosign}{\ensuremath{\text{How2Sign}}\xspace}
\newcommand{\csldaily}{\ensuremath{\text{CSL-Daily}}\xspace}
\newcommand{\phoenix}{\ensuremath{\text{Phoenix-2014T}}\xspace}

\newcommand{\dtwpa}{\ensuremath{\text{\textsc{DTW-PA-MPJPE}}}\xspace}

\newcommand{\bleufour}{\ensuremath{\text{BLEU-4}}\xspace}

\newcommand{\streamal}{\ensuremath{\text{Stream-AL}}\xspace}
\newcommand{\castreamal}{\ensuremath{\text{ca-Stream-AL}}\xspace}
\newcommand{\chunkwall}{\ensuremath{w_\text{chunk}}\xspace}

\newcommand{\bt}{\ensuremath{\text{back-translation}}\xspace}

\newcommand{\BT}{\ensuremath{\text{BT}}\xspace}

\newcommand{\sx}{\ensuremath{\mathbf{x}}}
\newcommand{\sy}{\ensuremath{\mathbf{y}}}
\newcommand{\Nx}{\ensuremath{|\mathbf{x}|}}
\newcommand{\Ny}{\ensuremath{|\mathbf{y}|}}
\newcommand{\xpref}[1]{\ensuremath{\mathbf{x}_{\le #1}}}
\newcommand{\ypref}[1]{\ensuremath{\mathbf{y}_{< #1}}}
\newcommand{\ptheta}{\ensuremath{p_{\boldsymbol{\theta}}}}
\newcommand{\Lpref}{\ensuremath{\mathcal{L}}}
\newcommand{\Tmax}{\ensuremath{T_\text{max}}}

\DeclareMathOperator{\argmax}{argmax}

\title{Simultaneous Translation between Sign Languages}

\author{
  Zetian Wu \quad Bowen Xie  \quad Stefan Lee \quad Liang Huang \\
  Oregon State University \\
  \texttt{\{wuzet, xiebo, leestef, liang.huang\}@oregonstate.edu}
}
\begin{document}

\maketitle

\begin{abstract}
Deaf and hard-of-hearing (DHH) signers cannot converse in real time across different sign languages today: existing sign-to-sign translation systems run offline, requiring the full source clip before any target sign is emitted.
Live use cases --- e.g.~broadcast interpretation and two-way video calls --- instead demand simultaneous output, while the source signer is still signing.
We present, to our knowledge, the first simultaneous \StoS\ (\StwoS) translation system, with two wait-$k$ regimes: test-time wait-$k$ inference applied directly to a full-sentence model, and a trained wait-$k$ model via stochastic multi-path supervision.
We further introduce \castreamal, a computation-aware latency metric for streaming output.
Averaged across six \StwoS\ directions on both a smaller human-verified test set and a larger synthetic \StwoS\ corpus, our streaming system achieves a $38\%$ \castreamal\ reduction while staying within a $9\%$~\dtwpa\ increase and a $2.1$ BLEU-4 drop compared to the full-sentence baseline.
A word-order case study probes how the streaming model handles word order mismatch between different sign languages --- a consequence of simultaneous translation.
\end{abstract}

\section{Introduction}
\label{sec:intro}

Sign language is the primary, most natural communication medium for many Deaf and hard-of-hearing (\dhh) users --- which is why major broadcasts such as White House press briefings provide live ASL interpretations alongside written captions, recognizing that the visual-spatial grammar of sign carries information that written text alone cannot.
Yet \dhh\ signers across different sign-language communities cannot converse in real time today. The \textit{cascade} approach, which chains a single sign-to-text (\ST) model with a spoken-language MT system and a text-to-sign (\TS) generator, incurs substantial latency; meanwhile, both the recent direct \StwoS\ model of~\citet{wu-etal-2026-direct} and the earlier approach of~\citet{inan-etal-2025-align} remain \emph{offline}, requiring the entire source clip before any target sign can be generated.
This rules out the live deployments --- e.g.~broadcast interpretation and two-way video calls --- where simultaneous \StwoS\ would matter most.

Simultaneous machine translation between spoken languages faces an analogous problem and commonly adopts \emph{wait-$k$} policies~\citep{ma-etal-2019-stacl}, which delay generation until $k$ source units have been observed and then proceed incrementally. We extend this principle to direct \StwoS\ translation as an end-to-end streaming pipeline, with the wait-$k$ schedule governing the entire path from source-sign encoding and LM inference to target-sign decoding and rendering. This setting also introduces a distinct latency challenge: unlike text tokens, generated sign chunks have non-negligible playback duration, during which new source input can arrive and subsequent chunks can be computed. Existing latency metrics therefore do not fully capture the timing experienced in streaming sign output.

We present, to our knowledge, the first simultaneous \StwoS\ translation system, adapting wait-$k$ under both test-time and trained regimes. We further introduce \castreamal, a source-centric, computation-aware latency metric designed for streaming sign generation, accounting jointly for source arrival, model computation, and target playback where AL~\citep{ma-etal-2019-stacl} and CA-AL~\citep{ma-etal-2020-simulmt} can underestimate latency.

Averaged across six \StwoS\ directions on both a small human-verified test set and a larger synthetic \StwoS corpus~\citep{wu-etal-2026-direct}, our system achieves a $38\%$ \castreamal\ reduction while staying within a $9\%$ mean per joint position error (temporally and spatially aligned using dynamic time wrapping and Procrustes-aligned, i.e.~\dtwpa)\citep{lin-etal-2023-one-stage, yu2024signavatars, shalev2022ham2pose, Baltatzis_2024_CVPR} increase and a $2.1$ BLEU-4 drop relative to the full-sentence baseline.

In summary, our contributions are:
\begin{itemize}[leftmargin=*,noitemsep,topsep=2pt]
  \item \textbf{First simultaneous direct \StoS\ translation (\S\ref{sec:method}):} we formalize the streaming-\StwoS\ task.
  \item \textbf{Two wait-$k$ regimes (\S\ref{sec:method}):} test-time inference on a full-sentence model and a trained wait-$k$ model via stochastic multi-path supervision.
  \item \textbf{Ca-\streamal latency metric (\S\ref{sec:metric}):} a computation-aware metric that captures fixed- duration emission missed by AL and CA-AL.
  \item \textbf{Word-order case study (\S\ref{sec:analysis}):} an analysis of how the streaming model handles word order mismatch between sign languages.
\end{itemize}

\section{Direct Sign-to-Sign Translation}
\label{sec:direct_s2s}

The direct \StwoS\ task takes a signed utterance in one sign language and produces a sign sequence in another within a single model, rather than cascading through sign-to-text recognition, spoken-language machine translation, and text-to-sign generation.
\citet{inan-etal-2025-align} present the first attempt in this setting and contribute methods for aligning multiple signed-language corpora to enable cross-lingual training; however, their generated signs achieve a back-translation BLEU-4 of 0 in two of the six translation directions, with the best direction reaching only 6.81, indicating the limited translation quality of this early system.
\citet{wu-etal-2026-direct} extend the line with two ingredients: (i)~a \bt-synthesized \SStoSS\ parallel corpus across \ASL\,/\,\CSL\,/\,\DGS, and (ii)~a single encoder--decoder architecture trained jointly on \TS\ and \StwoS, sharing a frozen VQ-VAE sign tokenizer with \soke~\citep{zuo-etal-2024-signs}.
Both prior systems are \emph{offline}: the encoder consumes the entire source clip before any target sign is emitted, which precludes live use.

\paragraph{Backbone we adopt.}
We follow the pipeline of~\citet{wu-etal-2026-direct} without architectural modification.
A frozen decoupled VQ-VAE~\citep{zuo-etal-2024-signs} tokenizes 25~Hz \smplx\ poses into three streams of sign tokens (body, left hand, right hand), with temporal downsampling factor 4 $\to$ effective 6.25~Hz token rate.
A single transformer, initialized from \mbart-large-cc25~\citep{liu-etal-2020-multilingual}, processes both \TS\ and \StwoS\ samples; sign-side input uses the \soke~\citep{zuo-etal-2024-signs} embedding fusion (one fused embedding per frame), text-side input uses standard subword embeddings.
Source\,/\,target sign language is signaled via special-token prefixes (\texttt{ASL}, \texttt{CSL}, \texttt{DGS}); the same module serves both \TS\ and \StwoS tasks.
At test time the decoder emits target sign tokens autoregressively; the VQ-VAE decoder reconstructs \smplx\ poses, which are rendered to an avatar in parallel with generation.

\section{Simultaneous Sign-to-Sign Translation}
\label{sec:method}

\begin{figure}[t]
  \centering
  \includegraphics[width=\columnwidth]{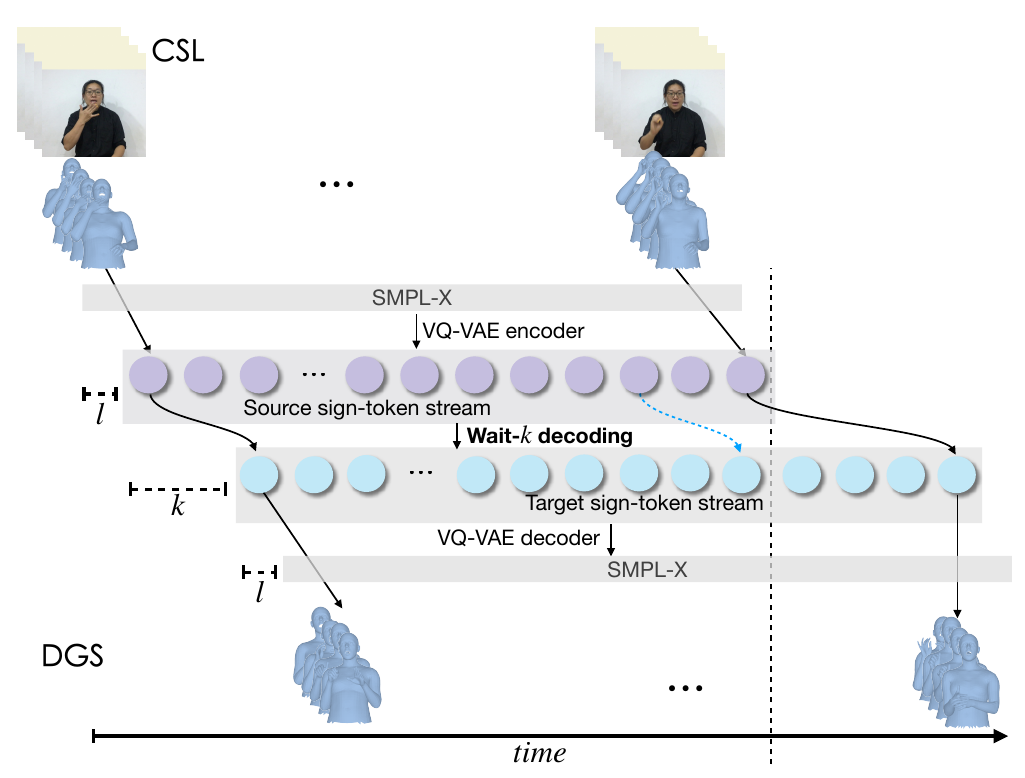}
  \caption{Simultaneous \StoS\ translation pipeline, shown for \CSL{}$\to$\DGS\ (the same architecture serves all six \StoS\ directions). The source video is incrementally fed to SMPL-X and tokenized by a VQ-VAE encoder into the source sign-token stream (purple, per-token duration $l$); wait-$k$ decoding (\S\ref{sec:ttwaitk}) emits target sign tokens (cyan, per-token duration $l$) after a $k$-token lag, and a VQ-VAE decoder maps them back to SMPL-X for rendering as the target (\DGS) avatar. We use $l{=}160$\,ms in this work (4 frames at 25\,fps). The vertical dashed line marks the end of source tokens at which target signing has already begun --- but under offline \StoS\ no target frame would be emitted before this point.}
  \label{fig:teaser}
\end{figure}

\subsection{Wait-$k$ Inference}
\label{sec:ttwaitk}

\paragraph{Policy.}
Let $\sx = (x_1, \dots, x_{\Nx})$ be the source sign-token stream and $\sy = (y_1, \dots, y_{\Ny})$ the target.
Each unit $x_i$ or $y_t$ is a sign token corresponding to 4 source/target frames at 25~fps (=160~ms of physical signing time). 
For full-sentence translation, the decoder conditions on the entire source:
\begin{equation}
y_t \;=\; \argmax_{y} \, \ptheta\!\left(\,y \;\middle|\; \sx,\; \ypref{t}\right).
\label{eq:full_decode}
\end{equation}

For simultaneous translation, let $g(t)$ denote the number of source tokens
available when generating the $t$-th target token. The decoder therefore
conditions only on the currently available source prefix:
\begin{equation}
y_t \;=\; \argmax_{y} \, \ptheta\!\left(
y \;\middle|\; \xpref{g(t)},\; \ypref{t}
\right).
\label{eq:decode}
\end{equation}
In our wait-$k$ setting, the read policy is
\begin{equation}
g_k(t) \;=\; \min(k+t-1,\, \Nx),
\label{eq:waitk_policy}
\end{equation}
which first reads $k$ source tokens and then reveals one additional source
token for each generated target token.
Once the source is exhausted, $g_k(t)=\Nx$ and decoding continues conditioned
on the full source.

\paragraph{Why re-encode per step.}
\mbart's encoder is bidirectional: for any observed source prefix, the hidden state $\mathbf{h}_i$ at position $i$ attends to all tokens in that prefix, including those after $i$. Consequently, when a new source token becomes available, the hidden states of earlier positions may also change and must be recomputed. A causal-mask approximation would discard the bidirectional inductive bias of the pre-trained backbone. We therefore interleave reading and writing: at each decoder step $t$ we reveal one more source token, re-run the encoder over the current prefix $\xpref{g_k(t)}$, and decode $y_t$ against the updated hidden states (Algorithm~\ref{alg:waitk}); $\Tmax$ caps the number of decoder steps. Per-step re-encoding scales encoder work as $\mathcal{O}(\Nx^3)$, but is acceptable in practice because end-to-end latency is currently dominated by the rendering stage (\S\ref{sec:metric}).

\begin{algorithm}[t]
\caption{Wait-$k$ decoding}
\label{alg:waitk}
\begin{algorithmic}[1]
\Require streaming source $\sx$ (final length $\Nx$), lag $k$, max target length $\Tmax$
\State $\sy \gets \langle\bos\rangle$ \Comment{$y_0=\langle\bos\rangle$}
\For{$t = 1, 2, \dots, \Tmax$}
    \State $g_k(t) \gets \min(k+t-1,\, \Nx)$ \Comment{read}
    \State $\mathbf{h} \gets \text{Encoder}(\xpref{g_k(t)})$ \Comment{re-encode}
    \State $y_t \gets \argmax_{y} \ptheta(y \mid \mathbf{h},\, \ypref{t})$ \Comment{Eq.~\ref{eq:decode}}
    \If{$y_t = \langle\eos\rangle$}
      \State \textbf{break}
    \EndIf
    \State {\bf yield} $y_t$  \Comment{incremental output}
    \State $\sy \gets \sy \circ y_t$
\EndFor

\end{algorithmic}
\end{algorithm}

\paragraph{Latency unit.}
A source-side unit of the transformer
is one sign token,
which we denote as a ``chunk''; 
each chunk includes 4 source frames, which
has length $w_\text{chunk} = 4/25$~s $= 160$~ms (25~fps source timeline).
AL is reported in \emph{seconds}.

\subsection{Trained Wait-$k$ Model}
\label{sec:trainwaitk}

Full-sentence training shows the model the full source at every target position; wait-$k$ inference (Eq.~\ref{eq:decode}) exposes position $t$ only to a prefix of $k+t-1$ source tokens.
The prefix-to-prefix objective of~\citet{ma-etal-2019-stacl} trains $\boldsymbol{\theta}$ under this wait-$k$ conditioning:
\begin{equation}
\Lpref(\boldsymbol{\theta}) \;=\; -\sum_{t=1}^{\Ny} \log \ptheta\!\left(\,y_t \;\middle|\; \xpref{g_k(t)},\; \ypref{t}\right).
\label{eq:pref}
\end{equation}
While Eq.~\ref{eq:pref} directly matches wait-$k$ inference, optimizing it exhaustively is impractical for sign-video inputs. Source sign sequences are much longer than typical text sequences, and each target position requires re-encoding a different source prefix, resulting in $\min(\Nx-k,\Ny)$ encoder forwards per sample. Batching these forwards exceeds GPU memory for typical sequence lengths. Following sampled-prefix training in simultaneous speech translation \citep{zhang-etal-2023-training}, we instead sample $m$ target positions uniformly and create $m$ replicas: replica $j$ receives $\xpref{k+t_j-1}$, with the loss masked at all positions except $t_j$.

\section {Computational-aware Lagging for Stream: \castreamal}
\label{sec:metric}

\begin{figure}[t]
  \centering
  \begin{subfigure}{\columnwidth}
    \centering
    \includegraphics[width=0.8\columnwidth]{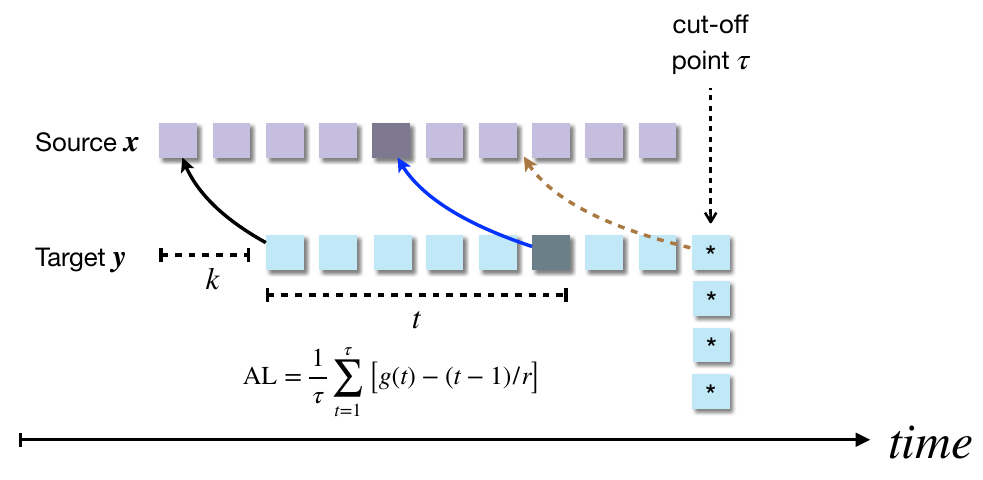}
    \caption{Classical AL: latency not counted for tail past cut-off point.}
    \label{fig:al-a}
  \end{subfigure}\\[0.0em]

  \noindent\makebox[\columnwidth][c]{
    \rule{\columnwidth}{0.5pt}
  }\\[0.0em]

  \begin{subfigure}{\columnwidth}
    \centering
    \includegraphics[width=0.8\columnwidth]{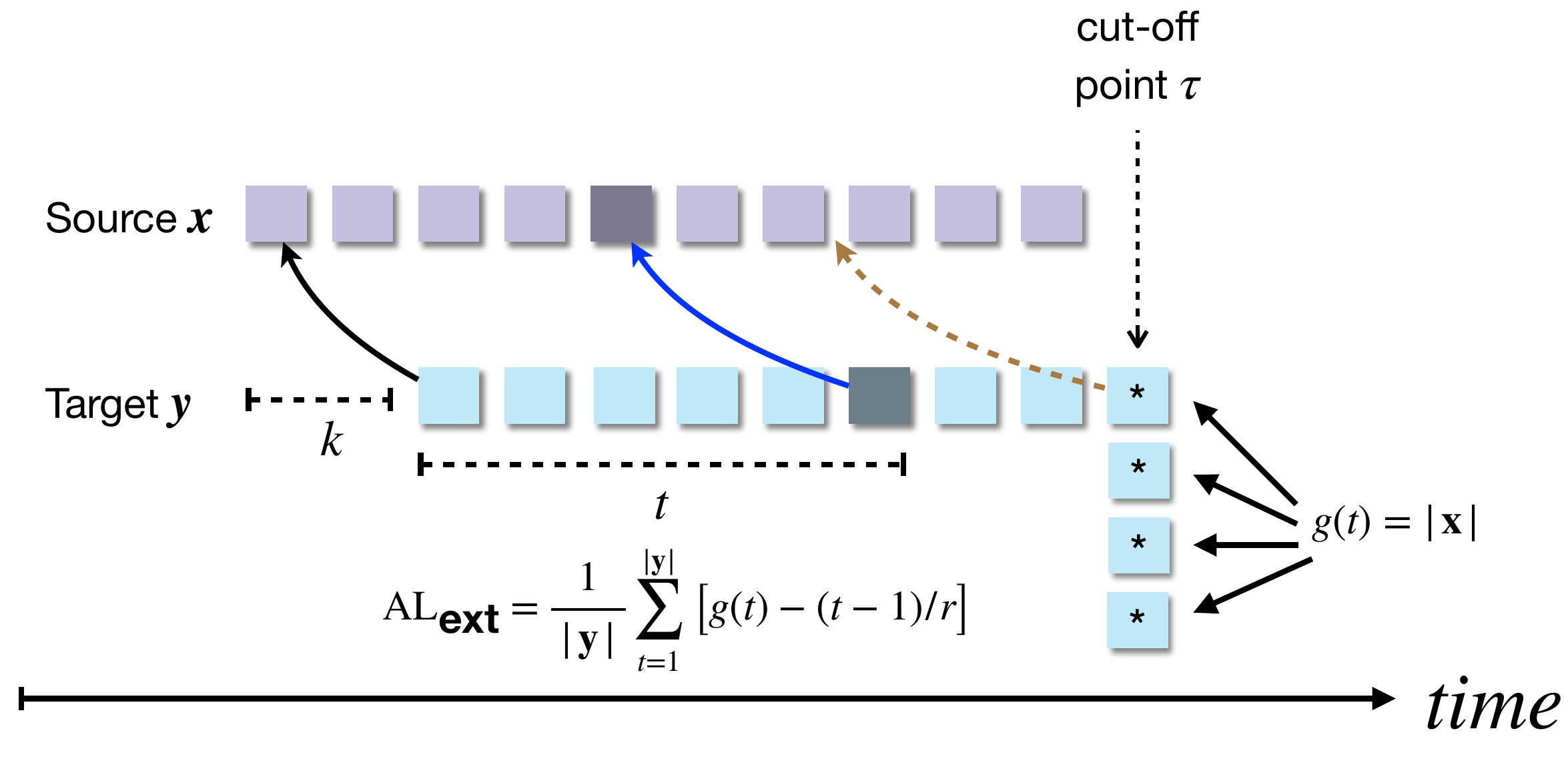}
    \caption{Fix to classical AL: extending the averaging
horizon from $\tau$ to the full target sequence.}
    \label{fig:al-b}
  \end{subfigure}\\[0.0em]

  \noindent\makebox[\columnwidth][c]{
    \rule{\columnwidth}{0.5pt}
  }\\[0.0em]
  
  \begin{subfigure}{\columnwidth}
    \centering
    \includegraphics[width=0.8\columnwidth]{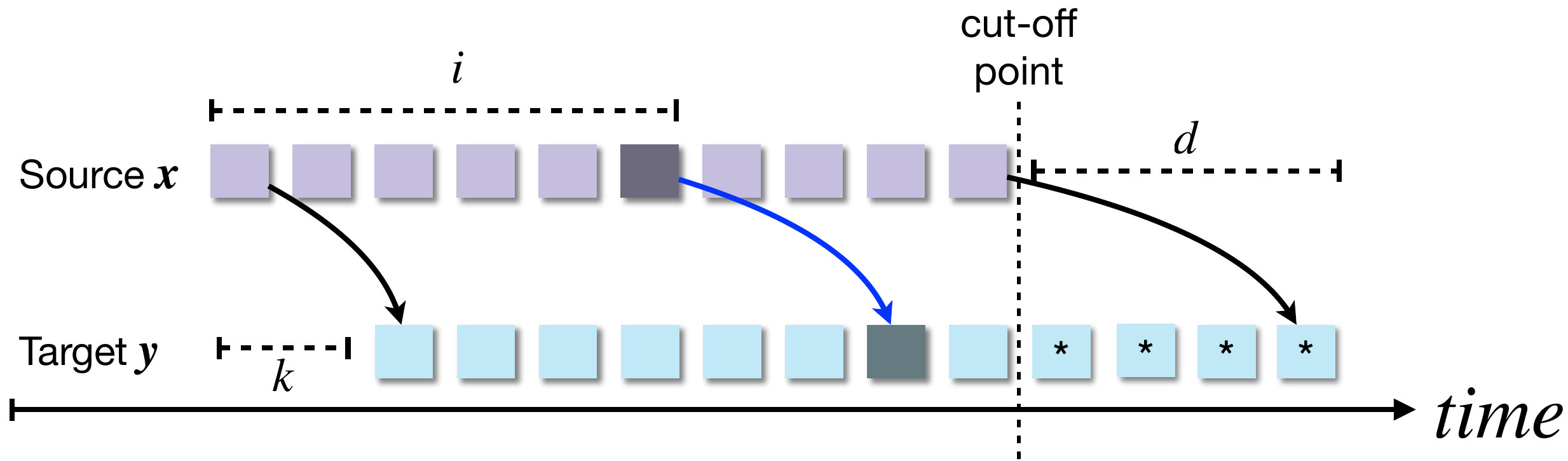}
    \caption{Our \streamal: shared $\chunkwall$-spaced time axis.}
    \label{fig:al-c}
  \end{subfigure}
  \caption{From classical AL to \streamal. \textbf{(a)} Dropping the $\tau$ cutoff captures the per-token theoretical lag of the target tail, but the metric remains target-centric and implicitly assumes the tail can be batch-emitted at the source-end instant $t = \Nx\chunkwall$ (visualized by the targets ``dropped down'' to a single column). \textbf{(b)} We fix the target-tail truncation issue by removing the source-end cutoff, which means extending the averaging horizon from the previous cutoff point $\tau$ to the full target sequence length $|\mathbf{y}|$. \textbf{(c)} \streamal\ lays source and target on a shared $\chunkwall{=}160$~ms time axis and sums the signed gap against the source-side proportional ideal. Every unit is a real $\chunkwall$ slot of wall-clock time, so the metric converts to seconds via $\times \chunkwall$.}
  \label{fig:al}
\end{figure}

We introduce computational-aware stream average lagging (\textbf{\castreamal}), a source-centric computation-aware latency metric for streaming sign output, reported in \emph{seconds} along the 25~fps source timeline (each emitted chunk spans $\chunkwall = 4/25$~s $= 160$~ms of wall-clock time). We begin by discussing the limitations of classical average lagging (AL), and develop \castreamal in two stages: a non-computation-aware formulation (\streamal) that establishes the source-centric view, followed by a computation-aware extension that incorporates per-step computation lag (\castreamal).

\subsection{Classical Average Lagging}
Classical average lagging, introduced by \citet{ma-etal-2019-stacl}, measures latency only over target steps up to the point at which the full source has been consumed. It is defined as
\begin{equation}
\mathrm{AL} = \frac{1}{\tau} \sum_{t=1}^{\tau} \big[g(t) - (t-1)/r\big]
\end{equation}
where $r=\frac{\Ny}{\Nx}$ is the target-to-source length ratio and $\tau=\min\{t:g(t)=\Nx\}$ is the first target step at which the full source has been consumed. 

\paragraph{Limitations}
This formulation has two limitations that motivate our extensions:
\begin{enumerate}
\item \textbf{Target-tail truncation.}
By definition, AL stops at $\tau$ and excludes all subsequent target steps from the latency computation (Fig.~\ref{fig:al-a}). These steps can still carry nonzero lag relative to the proportional ideal and should therefore contribute to the metric. When $r\approx1$, the omitted tail is typically short and the resulting error is small; as $r$ increases, however, the tail can account for a substantial fraction of the target sequence, making the underestimation increasingly severe.

\item \textbf{No duration for streamed target tails.}
The second issue arises for temporally realized outputs such as speech or sign video (i.e., in text-to-speech, speech-to-speech, or sign language translation scenarios). Unlike text, where the tail can be presented as a complete sequence once generated, a streamed output must unfold over time (at a reasonable rate) for the perceiver to consume it; its remaining tail therefore cannot be treated as if it were emitted instantaneously at the source end, nor can it be presented too fast which interferes with understanding. For example, in speech-to-speech translation, substantially increasing the playback rate degrades naturalness and can disrupt comprehension~\citep{zheng-etal-2020-fluent}.

\end{enumerate}

\paragraph{Removing the source-end cutoff.}
We first address the target-tail truncation by extending the averaging horizon from $\tau$ to the full target sequence:
\begin{equation}
\mathrm{AL}_{\mathrm{ext}} = 
\frac{1}{\Ny}
\sum_{t=1}^{\Ny}
\big[g(t)-(t-1)/r\big].
\label{eq:al-ext}
\end{equation}
Compared with classical AL, Eq.~\ref{eq:al-ext} retains the lag of every target step, including those emitted after the full source has been consumed (Fig.~\ref{fig:al-a}). This removes the truncation error, but does not address the second limitation: the formulation remains target-centric and does not account for the physical duration over which the target tail is streamed.

\subsection{\streamal}

To account for the streaming duration of the target tail, we reformulate lag on a shared source-centric timeline (Fig.~\ref{fig:al-c}). Rather than indexing latency by target emissions and asking how much source has been consumed, we index it by the fixed-rate source clock and ask how much target progress should have been completed at each source step. Since source chunks arrive every $\chunkwall$ seconds, each source index $i$ corresponds to a fixed wall-clock interval.

At source step $i$, the proportional ideal has reached target position $ir$, whereas the wait-$k$ schedule has reached $i-k$. We allow $i-k<0$ during the initial wait phase, interpreting it as a signed shortfall relative to the ideal rather than a physical number of emitted tokens. The gap $ir-(i-k)$ therefore measures the target-progress deficit on the shared streaming timeline. When the target progresses more slowly than the proportional ideal, this deficit accumulates over time and naturally reflects the unfinished target tail at the source end.

Averaging this signed gap over the source timeline gives
\begin{equation}
\begin{aligned}
\streamal
&=
\frac{1}{\Nx}
\sum_{i=1}^{\Nx}
\big[ir-(i-k)\big]
\\
&=
k+\frac{(\Nx+1)(r-1)}{2}.
\label{eq:streamal}
\end{aligned}
\end{equation}

\streamal\ is signed and may take negative values when $r<1$, indicating that the scheduled target progress is ahead of the proportional ideal. Since the gap in Eq.~\ref{eq:streamal} is measured in target-chunk units and each target chunk spans $\chunkwall$ seconds, multiplying by $\chunkwall$ converts \streamal\ to seconds. \streamal\ captures only the lag induced by the streaming schedule and assumes instantaneous computation; we next account for the additional wall-clock delay introduced by inference and rendering.

\begin{figure}[t]
  \centering
  \includegraphics[width=\columnwidth]{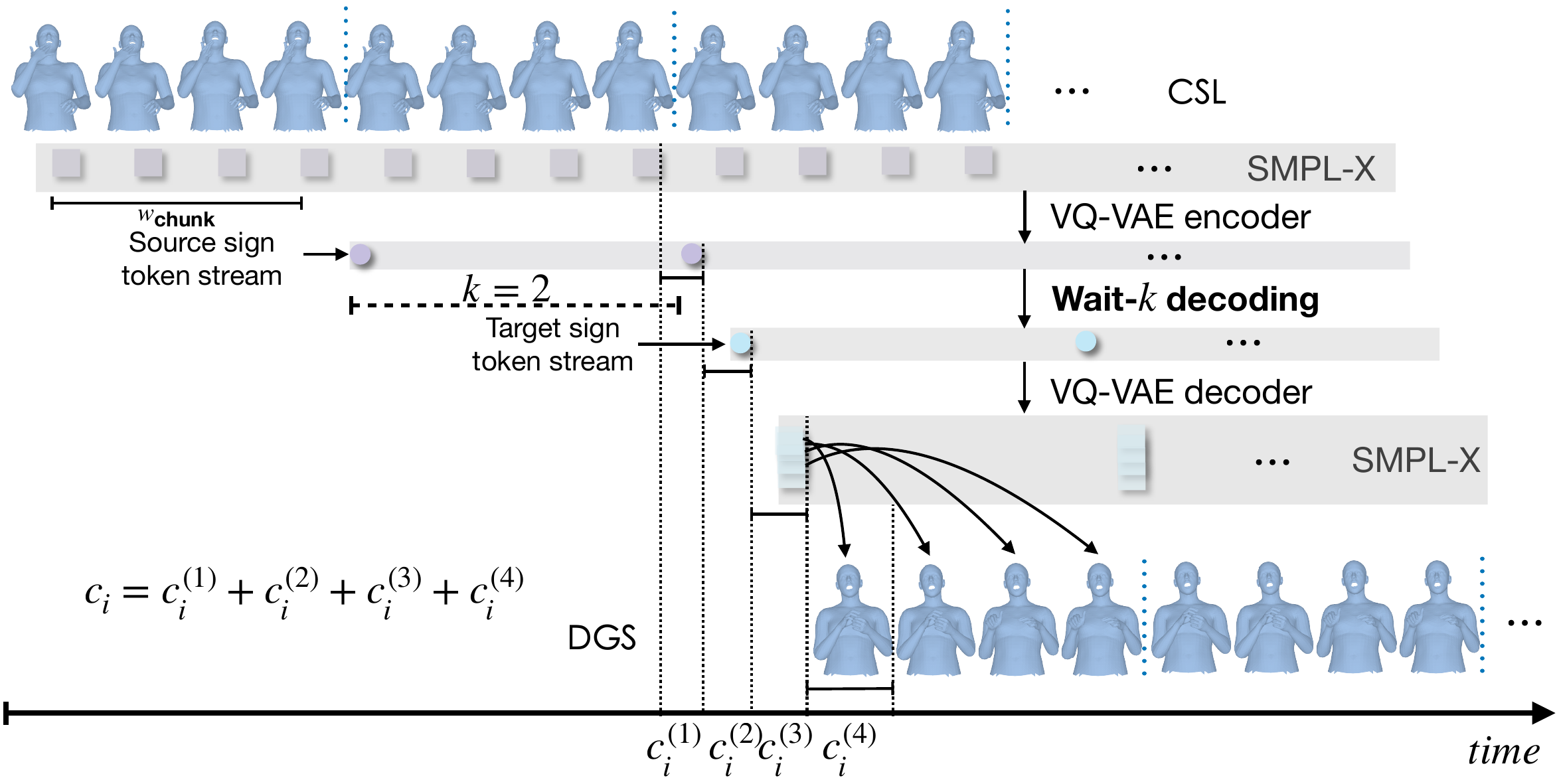}
  \caption{In \castreamal, for source step $i$, the chunk arrives at $i\times\chunkwall$, but the model completes generation at 
$t_i=\max(t_{i-1}, i\times\chunkwall)+c_i$, where $c_i$ includes VQ-VAE encoding($c_i^{(1)}$)/decoding($c_i^{(3)}$), transformer inference($c_i^{(2)}$), and SMPL-X rendering($c_i^{(4)}$). 
The wall-clock computation lag $\Delta_i=(t_i-i\times\chunkwall)/\chunkwall$ shifts the wait-$k$ schedule from $i-k$ to $i-k-\Delta_i$, yielding 
$\castreamal=\streamal+\bar{\Delta}$.}
  \label{fig:teaser-detailed}
\end{figure}

\subsection{Computation-aware extension.}
\streamal\ captures the lag induced by the streaming schedule but assumes instantaneous computation. In practice, encoding, transformer inference, decoding and rendering introduce additional wall-clock delay on the same shared timeline. We depict this extension in Fig.~\ref{fig:teaser-detailed}.

Let $c_i$ denote the computation incurred at source step $i$, including VQ-VAE encoding and decoding, transformer inference, and \smplx\ rendering. Let $t_i$ denote the wall-clock time at which step $i$ finishes. Computation for step $i$ can begin only after both source chunk $i$ has arrived and the previous step has completed, giving
\begin{equation}
t_i = \max(t_{i-1},, i\chunkwall) + c_i.
\label{eq:tmodel}
\end{equation}
Thus, any computation that exceeds the available time between source arrivals is carried forward as backlog.

We define the computation-induced delay at source step $i$ as
\begin{equation}
\Delta_i =\frac{t_i-i\chunkwall}{\chunkwall},
\end{equation}
expressed in chunk-duration units. This delay includes both the computation of the current step and any backlog accumulated from previous steps. Since each target chunk spans the same $\chunkwall$ seconds, a wall-clock delay of $\Delta_i\chunkwall$ corresponds to a backward displacement of $\Delta_i$ target-chunk positions on the shared timeline. The nominal wait-$k$ progress $i-k$ therefore becomes $i-k-\Delta_i$.

Substituting this computation-delayed progress into Eq.~\ref{eq:streamal} gives
\begin{equation}
\begin{aligned}
\castreamal
&=
\frac{1}{\Nx}
\sum_{i=1}^{\Nx}
\big[ir-(i-k-\Delta_i)\big]\\
&=
\streamal+\bar{\Delta},
\label{eq:castreamal}
\end{aligned}
\end{equation}
where $\bar{\Delta}=\frac{1}{\Nx}\sum_{i=1}^{\Nx}\Delta_i$ is the mean computation-induced delay. Thus, \castreamal\ decomposes into the schedule-induced lag measured by \streamal\ and the additional lag introduced by computation, and reduces to \streamal\ when $\Delta_i\equiv0$.

\paragraph{Pipeline boundary.}

The full pipeline is:
source video frames
$\to$ \smplx{} regression into source \smplx{} motion
$\to$ VQ-VAE encoding into source tokens
$\to$ transformer inference from source tokens to target tokens
$\to$ VQ-VAE decoding into target \smplx{} motion
$\to$ \smplx{} rendering into visible avatar output.
\castreamal measures latency from the source \smplx{} motion onward.
Accordingly, VQ-VAE encoding and decoding, transformer inference, and the downstream \smplx{}-to-avatar rendering stage are included in the per-step computation cost $c_i$.

The upstream regression from raw video to \smplx, however, is treated as an external preprocessing stage. We assume that it operates at or above the 25~fps source rate, so that it does not accumulate a growing backlog relative to the incoming stream. This assumption is practical with modern pose regressors, several of which approach or exceed real-time throughput on contemporary GPUs. For example, OSX~\citep{lin-etal-2023-one-stage} reports 12.2~fps on V100, which scales to approximately 56~fps on H100 under the reported $\sim$4.6$\times$ transformer-inference speedup, while faster alternatives such as SMPLer-X~\citep{cai-etal-2023-smpler} already exceed 25~fps on V100. Under this assumption, upstream regression contributes only a bounded processing offset rather than an accumulating delay, and is therefore excluded from \castreamal. If the regressor runs below the source frame rate, its backlog instead grows over time, in which case the reported \castreamal should be interpreted as a lower bound on the true end-to-end latency.

We retain the standard bidirectional encoder of~\citet{wu-etal-2026-direct}; under streaming inference, the available source prefix is re-encoded at each step, and this computation is included in $c_i$.

\paragraph{Relation to prior metrics.}
\streamal is related to duration-aware latency metrics such as ATD~\citep{kano-etal-2023-average}, but differs in its reference frame. ATD remains target-centric and measures delay with respect to corresponding input positions, whereas \streamal measures target progress on a shared source-time axis against the proportional ideal $ir$. This source-centric formulation accounts explicitly for the physical duration of the output stream and permits signed lag, including negative values when target progress is ahead of the proportional ideal.

The computation-aware extension is complementary to prior computation-aware AL formulations~\citep{ma-etal-2020-simulmt}. Rather than attaching computation cost only to output emission, \castreamal models the wall-clock completion time of every streaming step. The per-step cost $c_i$ therefore includes computation incurred even when no target token is emitted, such as repeated source-prefix encoding during the initial wait phase, together with decoding and rendering when output is produced. Through the recurrence in Eq.~\ref{eq:tmodel}, any computation that exceeds the available wall-clock budget is carried forward as backlog, allowing \castreamal to capture both per-step processing latency and its accumulation over the stream.

\section{Experiments}
\label{sec:experiments}

\subsection{Setup}
\label{sec:setup}

\paragraph{Backbone.}
\mbart-large-cc25 finetuned from the public pre-trained weights (\S\ref{sec:direct_s2s}); frozen VQ-VAE tokenizer from~\citet{zuo-etal-2024-signs}.
We follow the public training config of~\citet{wu-etal-2026-direct} (AdamW, lr $2{\times}10^{-4}$ cosine-annealed to $10^{-6}$, 150 epochs, effective batch size 256, 2$\times$H100).

\paragraph{Source frame rate.}
All three monolingual corpora (\howtosign, \csldaily, \phoenix) provide \smplx\ poses at \textbf{25~fps}, and we adopt this rate for the source sign streams throughout.
Combined with the VQ-VAE temporal downsampling factor of 4, one sign token spans 4 source frames $= 4/25 = 0.16$~s ($160$~ms); the effective token rate of the transformer is therefore $25/4 = 6.25$~Hz.
The \castreamal reported below are in seconds along this 25~fps source timeline.

\paragraph{Training corpus.}
A freshly generated back-translation(\BT) \StwoS\ corpus built following the \BT\ methodology of~\citet{wu-etal-2026-direct}.

\paragraph{Test sets.}
\textbf{BT:} held-out (text, sign) pairs run through our \BT\ pipeline; synthetic source, gold target.
This is distinct from the \texttt{BT-input} split of~\citet{wu-etal-2026-direct} --- same methodology, fresh built corpus.
\textbf{Strict:} the dataset from ~\citet{wu-etal-2026-direct} (\SStoSS pairs collected from \howtosign, \csldaily, \phoenix with alignment using their text translaition).

\paragraph{Directions.}
All six $(s \to s')$ with $s, s' \in \{\ASL, \CSL, \DGS\}$, $s \ne s'$.

\paragraph{Metrics.}
Quality: \dtwpa\ (primary, following~\citealp{wu-etal-2026-direct}), computed with the \texttt{fastdtw} algorithm~\citep{salvador-chan-2007-toward} at radius $r{=}2$; and \bleufour\ via the Sign Language Transformer \citep{camgoz-etal-2020-sign}.
Latency: \castreamal (\S\ref{sec:metric}).

\paragraph{Variants.}
\textbf{Full-sentence:} $k = \Nx$ baseline following~\citet{wu-etal-2026-direct}.
\textbf{TT-only:} wait-$k$ inference on the full-sentence checkpoint (\S\ref{sec:ttwaitk}); no retraining.
\textbf{Train+TT:} stochastic multi-path with $m{=}5$ (\S\ref{sec:trainwaitk}).\footnote{Teacher-forced \texttt{decoder\_input\_ids} are shifted from the unmasked labels and the position mask is applied afterward, keeping the gold prefix intact for the decoder. Non-\StwoS\ samples in the mixed \TS{}$+$\StwoS\ batches bypass replication.}

\paragraph{Latency sweep.}
$k \in \{1,\, 3,\, 5,\, 7,\, 9,\, 11,\, 13\}$ source sign tokens, i.e. $\approx \{0.16,\, 0.48,\, 0.80,\, 1.12,\, 1.44,\, 1.76,\, 2.08\}$~s.

\paragraph{Artifacts and licenses.}
All upstream artifacts are used under their respective public licenses. \mbart-large-cc25~\citep{liu-etal-2020-multilingual}, \texttt{pyrender}, and the \texttt{fastdtw} implementation~\citep{salvador-chan-2007-toward} are distributed under permissive open-source licenses; the decoupled VQ-VAE tokenizer~\citep{zuo-etal-2024-signs} and the OSX pose regressor~\citep{lin-etal-2023-one-stage} are released by their authors for research use. The three monolingual corpora (\howtosign, \csldaily, \phoenix) and the Strict subset (derived from~\citealp{inan-etal-2025-align} via~\citealp{wu-etal-2026-direct}) are distributed under research-only / non-commercial terms. Our use of every artifact --- as building blocks of a research system for sign-language translation --- is consistent with these intended-use conditions. We will release our \BT-synthesized \StwoS\ corpus and trained checkpoints under terms compatible with the strictest upstream license (research-only); downstream users should honor those restrictions, in particular the non-commercial / research-only terms of the source corpora.

\subsection{Quality--Latency Pareto Frontier}
\label{sec:pareto}

\begin{figure*}[t]
  \centering
  \begin{subfigure}[t]{0.48\textwidth}
    \centering
    \includegraphics[width=\linewidth]{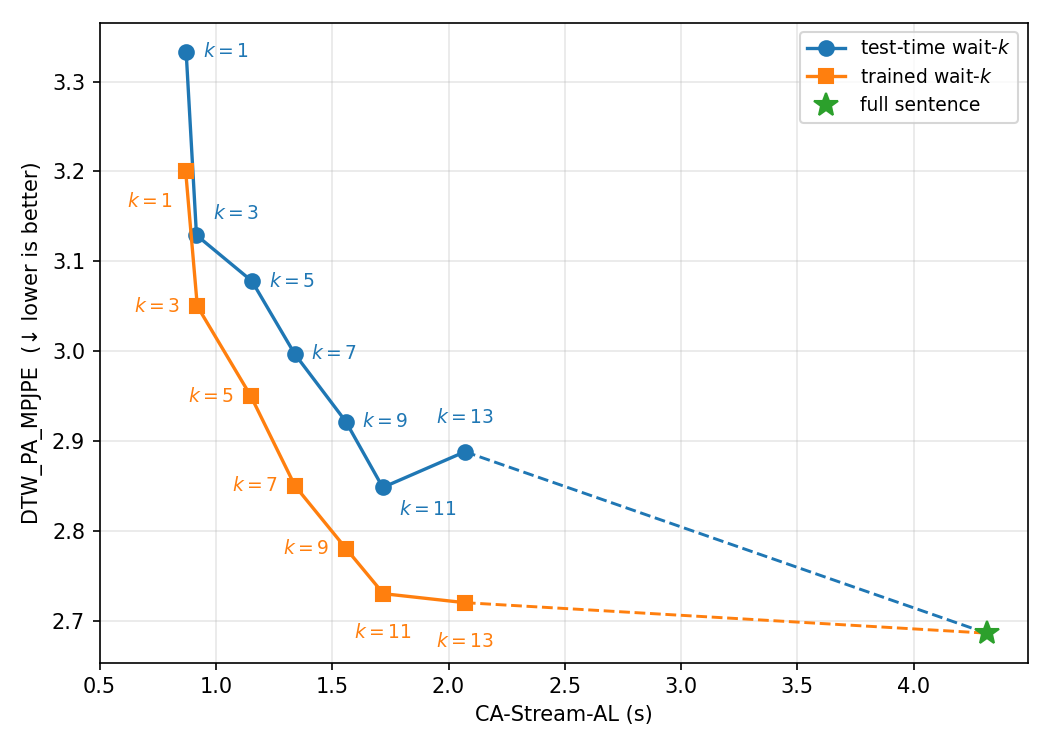}
    \caption{BT, \dtwpa\ ($\downarrow$)}
    \label{fig:pareto:bt_dtw}
  \end{subfigure}\hfill
  \begin{subfigure}[t]{0.48\textwidth}
    \centering
    \includegraphics[width=\linewidth]{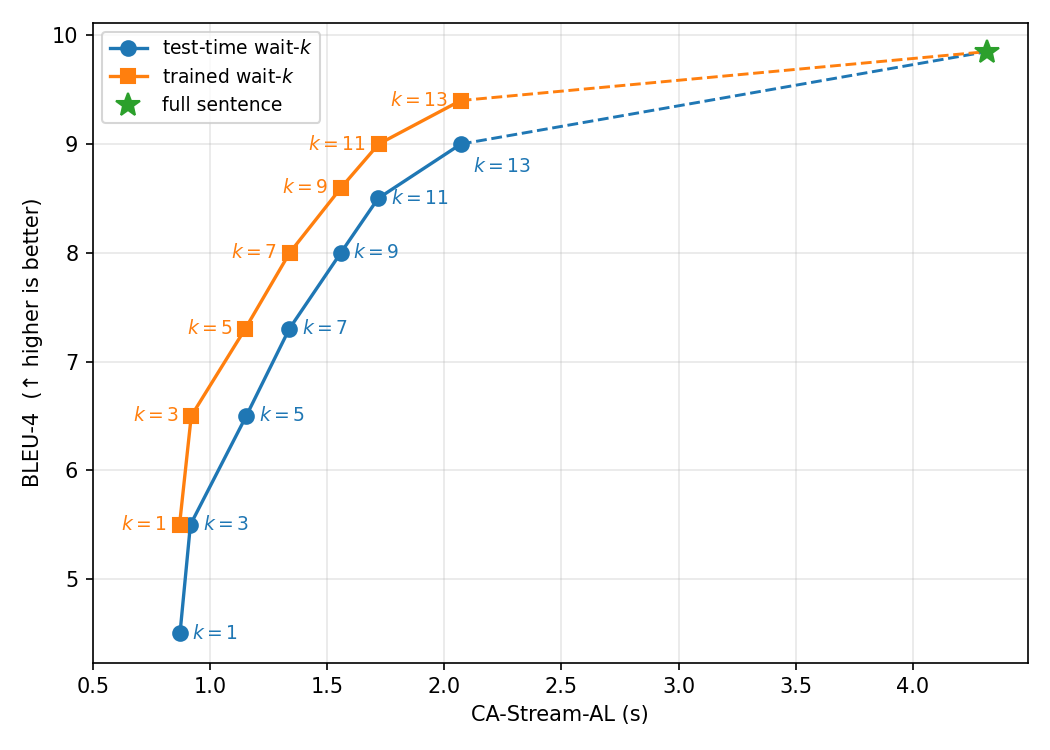}
    \caption{BT, \bleufour\ ($\uparrow$)}
    \label{fig:pareto:bt_bleu}
  \end{subfigure}\\[0.2em]
  \begin{subfigure}[t]{0.48\textwidth}
    \centering
    \includegraphics[width=\linewidth]{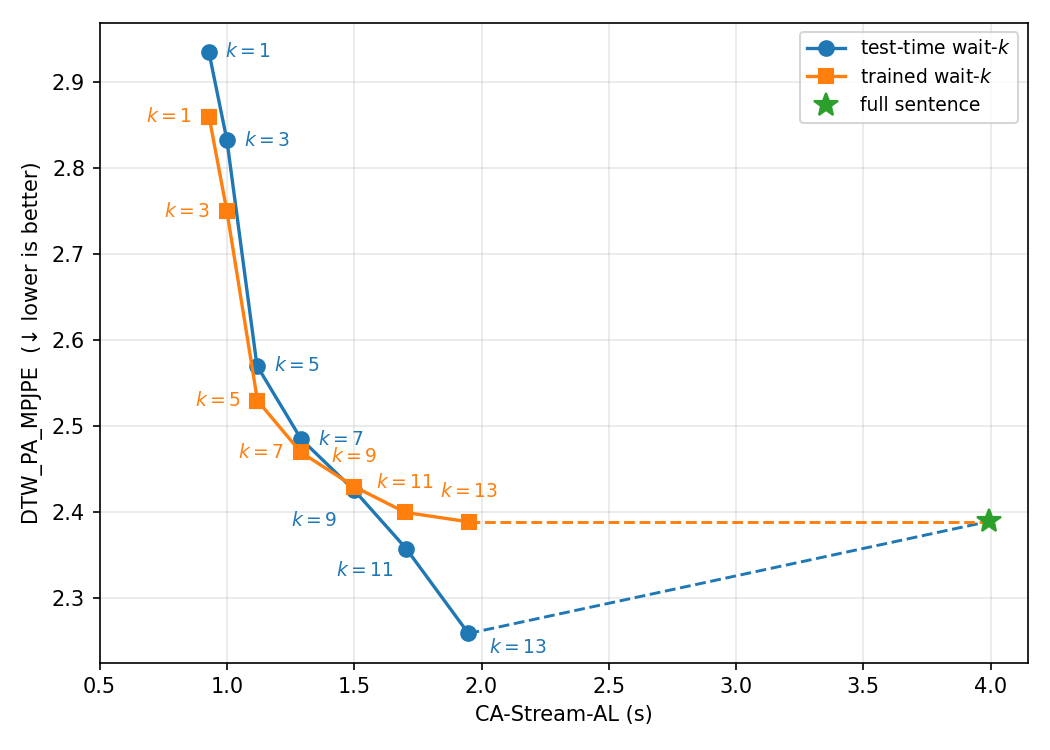}
    \caption{Strict, \dtwpa\ ($\downarrow$)}
    \label{fig:pareto:strict_dtw}
  \end{subfigure}\hfill
  \begin{subfigure}[t]{0.48\textwidth}
    \centering
    \includegraphics[width=\linewidth]{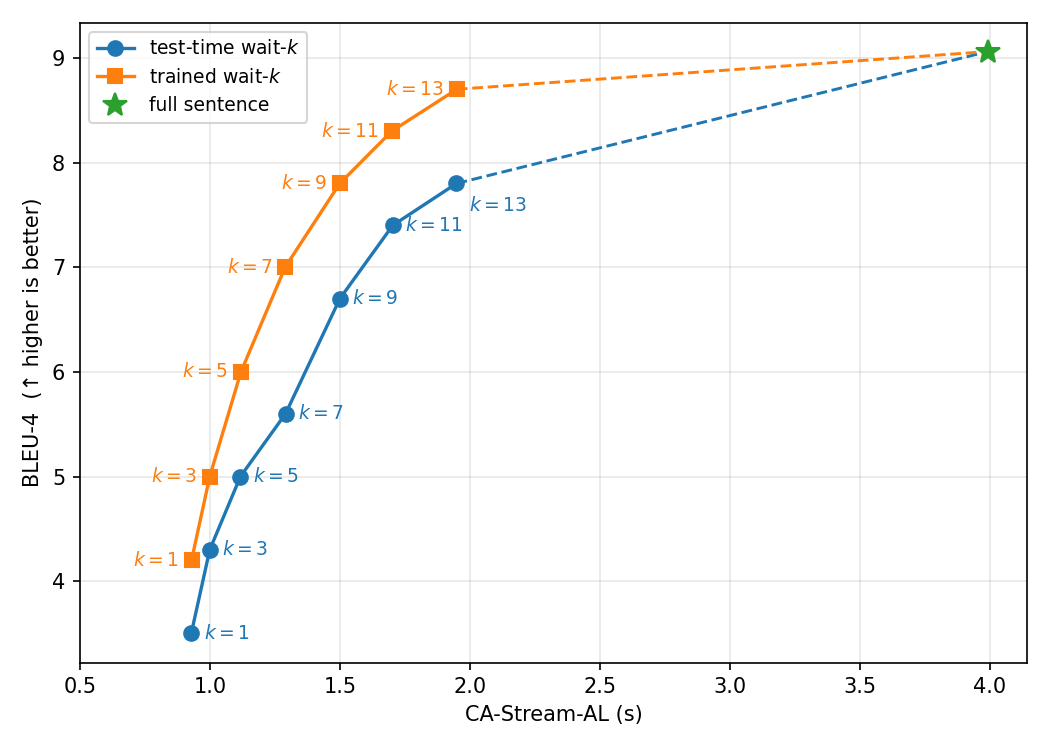}
    \caption{Strict, \bleufour\ ($\uparrow$)}
    \label{fig:pareto:strict_bleu}
  \end{subfigure}
  \caption{Quality--latency trade-off on \CSL{}$\to$\DGS, a representative direction out of the six \StoS\ directions we evaluate. Test-time wait-$k$ (blue) and trained wait-$k$ (orange) traced over $k \in \{1, 3, 5, 7, 9, 11, 13\}$; the full-sentence $k{=}\infty$ baseline (green star) is connected by a dashed segment. Rows: test set (BT, Strict); columns: quality metric (\dtwpa, \bleufour); x-axis: \castreamal\ (seconds).}
  \label{fig:pareto}
\end{figure*}

We sweep all six \StoS\ directions; Fig.~\ref{fig:pareto} shows \CSL{}$\to$\DGS as a representative case, comparing TT-only and Train+TT ($m{=}5$) over $k$ against the full-sentence $k{=}\Nx$ baseline in a 2$\times$2 grid (rows: test sets BT and Strict; columns: \dtwpa\ and \bleufour; x-axis: \castreamal).
Across all four panels, Train+TT dominates TT-only at matched \castreamal (except for a small portion in Fig.~\ref{fig:pareto:strict_dtw}), confirming that exposing the model to source-prefix supervision recovers quality that pure test-time truncation leaves on the table.
Concretely, at $k{=}7$ on \CSL{}$\to$\DGS\ Train+TT reaches \castreamal $= 1.34$\,s --- a $\sim 3\times$ reduction over the full-sentence checkpoint at $4.32$\,s --- trailing full-sentence \dtwpa\ only by $0.19$ and \bleufour\ by $2.4$.

\paragraph{Per-direction summary.}
Tab.~\ref{tab:bestk} and Tab.~\ref{tab:bestk-strict} report the best-tradeoff $k$ per direction on the BT and Strict test sets respectively, with the resulting (\castreamal, \dtwpa, \bleufour) under Train+TT, $m{=}5$.

\begin{table}[t]
  \centering
  \small
  \begin{tabular}{lrrrr}
    \hline
    Direction & $k^\star$ & AL\,(s)\,$_\downarrow$ & DTW\,$_\downarrow$ & BLEU\,$^\uparrow$ \\
    \hline
    \ASL{}$\to$\CSL & 7 & 1.97 & 3.74 & 7.5 \\
    \CSL{}$\to$\ASL & 9 & 3.71 & 4.79 & 10.4 \\
    \ASL{}$\to$\DGS & 7 & 1.15 & 3.85 & 7.4 \\
    \DGS{}$\to$\ASL & 9 & 2.41 & 3.55 & 9.0 \\
    \CSL{}$\to$\DGS & 7 & 1.34 & 2.88 & 7.5 \\
    \DGS{}$\to$\CSL & 9 & 3.05 & 3.27 & 7.4 \\
    \hline
  \end{tabular}
  \caption{Best-tradeoff $k^\star$ per direction on BT under Train+TT, $m{=}5$. AL\,=\,\castreamal\ (seconds); DTW\,=\,\dtwpa; BLEU\,=\,\bleufour\ from the Sign Language Transformer evaluator. AL admits possible negative values for directions with $r<1$.}
  \label{tab:bestk}
\end{table}

\begin{table}[t]
  \centering
  \small
  \begin{tabular}{lrrrr}
    \hline
    Direction & $k^\star$ & AL\,(s)\,$_\downarrow$ & DTW\,$_\downarrow$ & BLEU\,$^\uparrow$ \\
    \hline
    \ASL{}$\to$\CSL & 7 & 2.88 & 2.53 & 5.5 \\
    \CSL{}$\to$\ASL & 9 & 1.78 & 2.09 & 8.5 \\
    \ASL{}$\to$\DGS & 7 & 2.11 & 2.02 & 5.3 \\
    \DGS{}$\to$\ASL & 9 & 1.79 & 1.52 & 7.6 \\
    \CSL{}$\to$\DGS & 7 & 1.29 & 2.39 & 5.6 \\
    \DGS{}$\to$\CSL & 9 & 2.94 & 3.72 & 6.3 \\
    \hline
  \end{tabular}
  \caption{Best-tradeoff $k^\star$ per direction on Strict under Train+TT, $m{=}5$. Columns as in Tab.~\ref{tab:bestk}.}
  \label{tab:bestk-strict}
\end{table}

\section{Analysis: Word Order and Streaming Anticipation}
\label{sec:analysis}

Simultaneous translation surfaces a problem that the offline setting can ignore: the decoder cannot reorder content it has not yet observed.
At low $k$, the wait-$k$ policy can only \emph{wait} (paying latency) or \emph{anticipate} (risking error).
This tension is as sharp for sign-to-sign as for text-to-text simultaneous MT because the canonical word orders of \ASL, \CSL, and \DGS\ also differ substantially:
\ASL\ is broadly SVO with frequent topic-fronting;
\DGS\ is largely SOV, placing the verb at the clause end;
\CSL\ admits both SVO and SOV constructions and is topic-prominent, and --- unlike \DGS\ --- places sentential negation \emph{before} the predicate it scopes over.
A streaming \CSL{}$\to$\DGS\ system must therefore delay clause-final material across earlier source tokens and, in the negative case, promote the adjective ahead of the negator.

\begin{figure}[t]
  
  \raggedright
  \begin{subfigure}{\columnwidth}
    \centering
    \includegraphics[width=\linewidth]{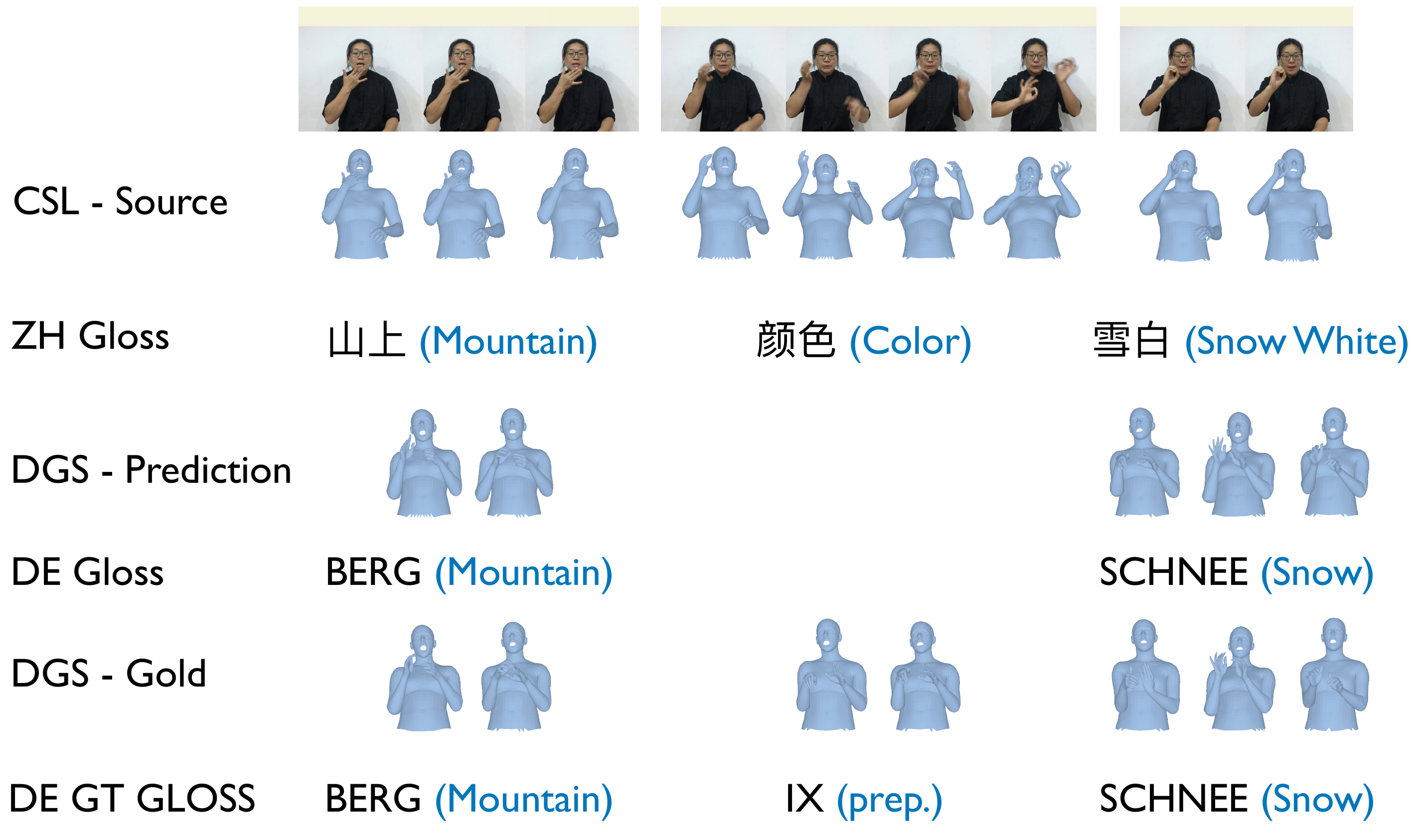}
    \caption{Order-aligned content: \CSL\ ``mountain / color / snow-white'' $\to$ \DGS\ \textsc{berg schnee}.}
    \label{fig:wordorder-good}
  \end{subfigure}\\[0.6em]

  \noindent\begin{subfigure}{\columnwidth}
    \raggedright
    \includegraphics[width=0.71\linewidth]{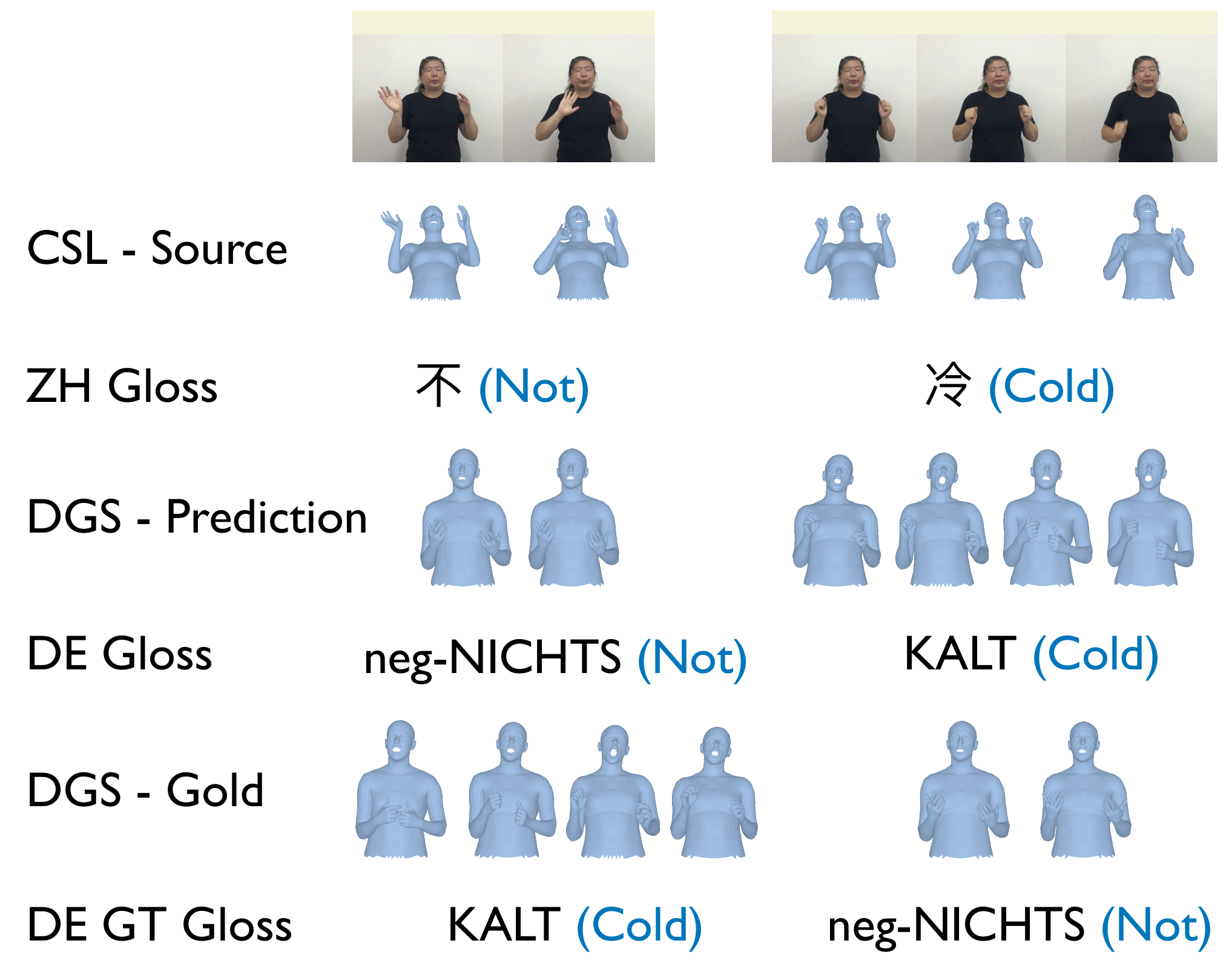}
    \caption{Reordered negation: \CSL\ \emph{neg}--adj $\to$ \DGS\ adj--\emph{neg}.}
    \label{fig:wordorder-bad}
  \end{subfigure}

  \caption{\CSL{}$\to$\DGS\ case study (Train+TT, $k{=}3$). \textbf{(a)}~When the canonical source and target orders align over content words, the model emits the right \DGS\ sequence and drops the \CSL\ topic-marker glossed ``color'' (\textit{yanse}) that has no \DGS\ counterpart. \textbf{(b)}~When the two languages disagree on the position of the negator, the streaming model emits source-order \textsc{neg-nichts kalt} as soon as the source negator (\textit{bu}, ``not'') arrives, instead of the canonical \DGS\ order \textsc{kalt neg-nichts}.}
  \label{fig:wordorder}
\end{figure}

We illustrate both outcomes with two \CSL{}$\to$\DGS\ picks from the Strict subset of~\citet{wu-etal-2026-direct}, decoded by Train+TT at $k{=}3$ (Fig.~\ref{fig:wordorder}).
In each panel, we show the source signing track, the gloss of each source segment, the model's predicted target signing, and the reference \DGS\ signing and gloss.

\paragraph{Good case (Fig.~\ref{fig:wordorder-good}).}
The \CSL\ source glosses to ``mountain / color / snow-white'' (\textit{sh\={a}nsh\`{a}ng / y\'{a}ns\`{e} / xuebai}), a topic-comment construction in which ``color'' is a discourse marker introducing the descriptor.
The reference \DGS\ gloss is \textsc{berg ix schnee} (mountain \textsc{ix} snow); the indexical \textsc{ix} has no lexical correspondent in the source.
The model emits \textsc{berg schnee}: it preserves the canonical mountain-then-snow order shared by both languages, decomposes the \CSL\ compound ``snow-white'' into the \DGS\ root \textsc{schnee}, and correctly drops the source-side ``color'' that does not surface in \DGS.
This is the regime where streaming costs little: anticipation is not required because the two canonical orders agree on the content sequence.

\paragraph{Bad case (Fig.~\ref{fig:wordorder-bad}).}
The \CSL\ source glosses to ``not cold'' (\textit{bu leng}), with the negator first.
The reference \DGS\ gloss is \textsc{kalt neg-nichts} (cold not): \DGS\ places sentential negation \emph{after} the predicate it scopes over.
A faithful streaming policy at $k{=}3$ would have to either wait for the adjective (``cold'') to arrive before committing to anything --- foregoing the latency advantage of streaming --- or anticipate the adjective from a single negation sign, which is under-determined. (Noted: To avoid confusion, $k$ in this paper always refers to sign tokens rather than the gloss unit. So wait-$3$ means waiting for three sign tokens when we can not even get the meaning of "cold" at the beginning.)
The model takes neither path: it emits \textsc{neg-nichts kalt} the moment the negator is observed, copying source order and violating the \DGS\ canonical order on a two-token utterance.
The miss is exactly the failure mode \citet{ma-etal-2019-stacl} predict for short clauses whose reordering window exceeds $k$.

\paragraph{Takeaway.}
The two picks bracket the regime: when source and target canonical orders agree over content (Fig.~\ref{fig:wordorder-good}), small-$k$ streaming reproduces the target sequence and even resolves source-only function words; when they disagree on a short, clause-internal swap (Fig.~\ref{fig:wordorder-bad}), the wait/anticipate tradeoff is unfavorable at $k{=}3$ and the model defaults to source order.
A quantitative version of this analysis --- per-direction anticipation lead time on a gloss-tagged subset --- is left to future work and noted in Limitations.

\section{Related Work}
\label{sec:related}

\paragraph{Simultaneous machine translation.}
We adopt the \emph{prefix-to-prefix} framework of~\citet{ma-etal-2019-stacl}, which factorizes streaming generation as $p(y_t \mid x_{\le g(t)},\, y_{<t})$ for a monotonic schedule $g$.
The wait-$k$ policy is the special case $g(t) = k + t - 1$.
AL~\citep{ma-etal-2019-stacl}, AP~\citep{cho-esipova-2016-can}, and DAL~\citep{cherry-foster-2019-thinking} provide the classical text-MT latency framework; we follow~\citet{ma-etal-2019-stacl} in using greedy decoding (beam search is ill-defined under partial-source observation).
\citet{ma-etal-2019-stacl} is text-to-text; we generalize the framework to sign-token chunks on a multilingual sign backbone, introduce \castreamal (\S\ref{sec:metric}) for streaming output in which each emitted chunk has a fixed wall-clock duration, and contribute a stochastic multi-path training objective (\S\ref{sec:trainwaitk}) that makes prefix-to-prefix tractable on the bidirectional encoder of the underlying direct-\StwoS\ model.
\citet{elbayad-etal-2020-efficient} train a single model that serves multiple $k$ at inference, sidestepping the per-$k$ retraining cost discussed in our Limitations.

\paragraph{Duration-aware latency metrics.}
\citet{kano-etal-2023-average} introduces ATD, which considers target-token duration in delay accounting and is the closest prior metric in spirit to \castreamal.
ATD is target-centric (per-target-token delay against the corresponding clipped input index) and conventionally non-negative; \castreamal is source-centric, uses the proportional diagonal $i\times r$ ($r=\frac{\Ny}{\Nx}$) as reference, admits a closed form for wait-$k$, and permits negative values when $r<1$.
\citet{ma-etal-2020-simulmt} extends wait-$k$ from text-to-text to streaming speech-to-text translation and introduces computation-aware AL (CA-AL); \castreamal applies the same computation-aware idea to our source-centric formulation.

\paragraph{Streaming sign-language translation (sign $\to$ text).}
\citet{yin-etal-2021-simulslt} introduce SimulSLT, which applies wait-$k$ to full-sentence sign-to-text translation~\citep{camgoz-etal-2018-neural,camgoz-etal-2020-sign} with a learned gloss-boundary predictor that decides when to read versus emit; their target is spoken-language text, and latency is measured by classical (target-centric) AL on text tokens.
Our setting differs along two axes: the target is rendered sign output emitted in fixed wall-clock chunks rather than text tokens at an instant --- which is what motivates the source-centric, computation-aware \castreamal\ (\S\ref{sec:metric}) --- and the source itself is a sign stream, making the present work, to our knowledge, the first streaming \StoS\ system in the literature.

\paragraph{Text $\to$ sign (\TS) translation.}
\TS translation (or in some literature called sign language production) generates sign output from spoken-language text, either as continuous pose sequences with progressive Transformers~\citep{saunders-etal-2020-progressive} or via motion-graph and GAN-based pipelines~\citep{stoll-etal-2020-text2sign}.
\TS translation is offline and consumes text; we generate sign tokens conditioned on a sign-language source under a streaming policy, inheriting the prefix-to-prefix latency budget rather than full-source access.

\section{Conclusion}
\label{sec:conclusion}

We presented, to our knowledge, the first simultaneous \StwoS\ translation system, closing the gap between offline \StwoS\ models and the live deployments --- broadcast interpretation, two-way video calls --- that require target signing to begin while the source signer is still signing.
We adapted the wait-$k$ policy to sign-token streams in two regimes --- test-time inference on a full-sentence backbone and a trained wait-$k$ model via stochastic multi-path supervision --- and introduced \castreamal, a source-centric computation-aware latency metric that accounts for the fixed wall-clock duration of each emitted sign chunk, which target-centric AL and CA-AL under-report.
Averaged across the six \StwoS\ directions, our streaming system cuts \castreamal\ by $38\%$ while staying within a $9\%$ \dtwpa\ increase and a $2.1$ \bleufour\ drop relative to the full-sentence baseline.
A word-order case study surfaced the central new challenge of going simultaneous: the decoder cannot reorder content it has not yet observed, so at low $k$ it must trade latency for anticipation on directions whose canonical orders disagree.
We hope \castreamal\ and the streaming-\StwoS\ task spur further work toward real-time cross-lingual sign communication.

\section*{Limitations}
\label{sec:limitations}

\paragraph{Length-ratio-agnostic schedule.}
Both wait-$k$ regimes (\S\ref{sec:method}) use the fixed unit-slope schedule $g(t){=}k{+}t{-}1$: after the initial $k$-token wait, exactly one source token is revealed per emitted target.
The six directions, however, have source/target length ratios far from $1$ ($r=\frac{\Ny}{\Nx}$ from $\approx 0.52$ to $\approx 2.44$; \S\ref{sec:ttwaitk}), so a unit-slope read/write schedule drifts linearly from the proportional ideal $i\times r$ --- the $\frac{(\Nx+1)(r-1)}{2}$ term that dominates \streamal\ (Eq.~\ref{eq:streamal}) when $r$ is far from $1$.
A length-ratio-aware \emph{catch-up} policy~\citep{ma-etal-2019-stacl} would instead match the read:write rate to the empirical ratio: for a direction whose source runs $\approx 1.2\times$ the target ($\approx 5$ target tokens per $6$ source tokens), it reveals two source tokens before one of every six writes --- e.g.\ the fifth and sixth source tokens are read together --- keeping the schedule parallel to the slope-$r$ ideal rather than slope $1$, and symmetrically writes two targets per read when the target is the longer side.
We use the ratio-agnostic schedule for simplicity and direct comparability with text-MT wait-$k$; biasing the schedule by the measured per-direction ratio is left to future work and would tighten \streamal\ most on the highly asymmetric directions.

\paragraph{Single-$k$ checkpoint.}
Each Train+TT model commits to one value of $k$ at training time; serving a different latency budget at inference requires retraining.
Multi-$k$ schemes~\citep{elbayad-etal-2020-efficient} sidestep this but at additional implementation cost --- left to future work.

\paragraph{Stochastic multi-path hyperparameters.}
We use uniform position sampling with $m{=}5$ replicas for the stochastic multi-path objective (\S\ref{sec:trainwaitk}).
Neither $m$ nor the sampling distribution (e.g., geometric, early-position-biased, schedule-dependent) was tuned; both could shift the early-segment quality at low $k$.

\paragraph{Synthetic training data.}
Train+TT is supervised on back-translation (\BT) synthetic \StwoS\ pairs (\S\ref{sec:setup}): the source side is generated by the \TS\ direction of the same backbone and so inherits its biases.
Whether these biases interact with the streaming objective --- e.g., by encouraging source-order anticipation when the target order would differ --- has not been characterized.

\paragraph{Qualitative anticipation analysis.}
\S\ref{sec:analysis} inspects a small number of cases.
A quantitative anticipation metric (e.g., source-aligned verb-emission lead time on a gloss-tagged subset) is left to future work.

\paragraph{No human evaluation.}
Quality is reported via \dtwpa\ and a downstream-evaluator \bleufour; native-signer human evaluation of streamed output is left to future work and would be the right tool for assessing perceived choppiness or unnatural early-segment artefacts.

\paragraph{Rendering bottleneck.}
Our \smplx{}-to-avatar rendering pipeline (\texttt{pyrender}) runs at $\approx 30$~fps, contributing $\approx 132$~ms per output chunk to \castreamal --- by far the dominant per-output cost, against $\sim 10$~ms for the encoder forward and LM decoder step combined.
Production-grade real-time avatar renderers (game-engine or neural-rendering systems) run substantially faster; swapping in such a renderer would tighten \castreamal\ numerically without changing the streaming policy.
A render-aware system design (e.g., overlapped render and decode) is outside this paper's scope.

\paragraph{Per-step re-encoding cost.}
Schedule construction in \S\ref{sec:ttwaitk} re-runs the bidirectional encoder on each successive source prefix and so scales encoder work as $\mathcal{O}(\Nx^3)$ (matching \S\ref{sec:ttwaitk}).
This is tolerable today because rendering dominates the per-chunk cost, but the constraint binds once rendering speeds up; a causal-mask approximation would amortize the cost at the price of discarding the bidirectional prior of the pre-trained backbone.

\paragraph{\smplx\ input assumption.}
We assume the upstream raw-video~$\to$~\smplx\ pose-estimation front-end runs at $\ge 25$~fps (matching the source frame rate); under this assumption extraction is a sub-chunk additive offset (\S\ref{sec:metric}), but below 25~fps a backlog accumulates linearly with $\Nx$ and our \castreamal\ becomes a lower bound on real user-perceived latency.
The OSX regressor~\citep{lin-etal-2023-one-stage} we adopt and faster alternatives such as the SMPLer-X family~\citep{cai-etal-2023-smpler} comfortably clear 25~fps on modern GPUs (\S\ref{sec:metric}), but a sustained deployment over long inputs would need to verify this empirically.

\paragraph{Greedy decoding.}
Following~\citet{ma-etal-2019-stacl}, we use greedy decoding because beam search is ill-defined under partial-source observation.
This leaves quality on the table relative to the (non-streaming) beam-search regime and is a shared limitation of the simultaneous-MT literature rather than specific to our work.

\paragraph{No external streaming-\StwoS\ baseline.}
To our knowledge there is no published simultaneous \StwoS\ system; we therefore compare only against our own full-sentence backbone and the test-time wait-$k$ inference variant.
Stronger external comparison will be possible once additional simultaneous \StwoS\ systems appear.

\paragraph{Coverage.}
Three sign languages (\ASL\,/\,\CSL\,/\,\DGS); minority sign languages are not represented.
Streaming systems should be deployed with care for community-specific norms.

\bibliography{custom}

\end{document}